\documentclass{article}

\usepackage{microtype}
\usepackage{graphicx}
\usepackage{subfigure}
\usepackage{lipsum}
\usepackage{booktabs} 

\usepackage{hyperref}
\hypersetup{
	pdfcreator = {},
	pdfproducer = {}
}

\usepackage[accepted]{icml2019}

\usepackage{amsmath}
\usepackage{amssymb}
\usepackage{enumitem}
\setlist{nolistsep}

\icmltitlerunning{Same, Same But Different - Recovering Neural Network Quantization Error Through Weight Factorization}

\begin{document}
	\twocolumn[

	\icmltitle{Same, Same But Different - Recovering Neural Network Quantization Error Through Weight Factorization}
	
	
	
	\icmlsetsymbol{equal}{*}
	
	\begin{icmlauthorlist}
		\icmlauthor{Eldad Meller}{}
		\icmlauthor{Alexander Finkelstein}{}
		\icmlauthor{Uri Almog}{}
		\icmlauthor{Mark Grobman}{}
		
	\end{icmlauthorlist}
	
	
	
	\icmlkeywords{Machine Learning, ICML}
	
	\vskip 0.15in

	
	
	
%
%
	\begin{center}
		{
			\textbf{Hailo Technologies}
%
		}
		\vskip 0.15in
		\end{center}
	
		]

	\begin{abstract}
		Quantization of neural networks has become common practice, driven by the need for efficient implementations of deep neural networks on embedded devices. In this paper, we exploit an oft-overlooked degree of freedom in most networks - for a given layer, individual output channels can be scaled by any factor provided that the corresponding weights of the next layer are inversely scaled. Therefore, a given network has many factorizations which change the weights of the network without changing its function. We present a conceptually simple and easy to implement method that uses this property and show that proper factorizations significantly decrease the degradation caused by quantization. We show improvement on a wide variety of networks and achieve state-of-the-art degradation results for MobileNets. While our focus is on quantization, this type of factorization is applicable to other domains such as network-pruning, neural nets regularization and network interpretability.

	\end{abstract}

	\section{Introduction} \label{sec:Introduction}
	Early efforts in the field of deep learning have focused mostly on the training aspect of neural networks. The success of these efforts has led to widespread deployment of trained neural networks in data-centers \cite{Park2018b,Jouppi2017} and on embedded devices \cite{Ignatov2018} where they are used for inference which in turn emphasized the need to make the inference phase more efficient. Quantization, which means conversion of the arithmetic used within the net from high-precision floating-points to low-precision integers, is an essential step for efficient deployment, however, quantization degrades network performance. Here, we follow the commonly used quantization scheme described in \citet{Jacob2017} but note that other schemes exist \cite{Vanhoucke2011,Gupta2015,Courbariaux2014} to which our results apply as well. Briefly, integer quantization consists of approximating real values with intergers according to $ x_Q=x/scale $ where $scale=(max(x)-min(x))/2^N $ and N is the number of bits used in the approximation. Each layer's weights and activations are given a different scale according to their extremum values. The noise introduced by this limited precision approximation encapsulates a fundamental dynamic range-precision trade-off.

	Existing approaches to decreasing induced degradation are ‘quantization-aware’ training \cite{Jacob2017,Banner2018a,Zhou2017,Zhou_2018_CVPR,McKinstry2018} and reducing the dynamic range of activations by clipping outliers \cite{Migacz2017,Choi2018,Banner2018b}. Training is a powerful method but it is time-consuming, hard to implement, and requires access to the original training dataset which might not always be available (e.g. when the user wishes to use an off-the-shelf pre-trained model). Clipping has limited effect since it only addresses noise from activation quantization. 	
	
	In this paper, we propose a different approach. Instead of focusing on improving the quantization process itself, we explore equivalent weight arrangement that make the net less sensitive to quantization. An equivalent weight arrangements is a factorization that changes the weights of the networks without changing its function - i.e. for a given input, the network output remains the same. During quantization, the range of each layer is set by the channel with the largest absolute activation which we term the dominant channel. This single channel determines the noise in all other channels, many of which have smaller values. Therefore, amplifying these channels to match the dominant channel while compensating the change at the next layer can reduce the effect of the noise.

	We begin by analyzing the noise introduced by quantization of weights and activations in terms of signal-to-quantization-noise ratio (SQNR). We inspect the effect of channel-scaling on SQNR and introduce an equalization procedure which, under some constraints, tries to  scale each output channel such that its range matches that of the dominant channel. We show that equalization reduces the layer SQNR and then apply equalization iteratively layer-by-layer and empirically show that the overall post-quantization degradation of the network decreases. Since our approach can be a pre-processing step prior to quantization, it is fully compatible with other approaches that improve the quantization process. Nevertheless, an appealing aspect of our scheme is that for most nets it reduces the quantization induced degradation enough as to make quantization-aware training unnecessary and thus facilitates rapid deployment of quantized models.
	
	The main contributions of this paper are:
	\begin{itemize}
		\setlength\itemsep{1em}
		\item \textbf{Inversely proportional factorization}: we show the utility of weight factorization for the task of quantization. Future work can benefit by exploiting these factorizations in other settings. To the best of our knowledge this work is the first to both highlight and show the usefulness of inversely proportional factorizations.
		\item \textbf{Equalization}: we show that layers which have channels with similar ranges are less affected by quantization and we show how to transform a network closer to this ideal. We also perform a quantitative analysis of the effect of equalization on quantization noise and quantization induced degradation for a wide range of network architectures. 
	\end{itemize}

	\section{Previous Work} \label{sec:Previous Work}
	\textbf{Equalization.} Having channels with similar dynamic ranges motivated \cite{Jacob2017} to use of Relu6 activations which were subsequently used in MobileNets \cite{Howard2017}. However, in practice, many channels remain un-clamped and the dynamic range strongly varies within a layer \cite{Sheng2018}. It was also observed \cite{Krishnamoorthi2018} that having a scale for each channel of a layer greatly improves quantization performance. While effective for networks where most of the degradation stems from the quantization of weights, it doesn't improve performance of networks that are degraded by the quantization of activation such as DenseNets \cite{Huang2016}.

	\textbf{Quantization Noise Analysis.} The properties of noise induced by the quantization of both activations and weights were analyzed in \citet{Lin2015} focusing on the optimal bit width assignment to each layer across the network. We follow a similar analysis but focus on the dynamic ranges of individual channels within a layer. \citet{pmlr-v70-sakr17a} gives an upper bound on the relationship between SQNR and network accuracy. An empirical disambiguation of the contributions of activation and weight noise to total degradation was given in \citet{Krishnamoorthi2018} for several networks. The consensus of previous works seems to be that weight quantization is responsible for the bulk of degradation but we show the opposite for some common networks.

	\section{Theoretical Foundation} \label{sec:Theoretical Foundation}
	For a given network architecture there exist many weight assignments that result in networks that realize the same mapping from input to output. Thus, we are afforded with an important degree of freedom enabling us to choose assignments that have desirable properties for the task at hand.
	We show, that for a family of networks it is possible to gradually switch between equivalent assignments through the use of inversely proportional factorizations. These factorizations enable us to scale individual channels within a layer by any positive factor. We then analyze the source of quantization noise and show that by scaling channels we can improve the SQNR withing a layer. Our analysis is done for Convolutional Neural Networks (CNNs) but the same principles can apply to other types of nets as well. Since our focus is on trained networks we assume that batch-normalization \cite{IoffeS15} layers are always folded back to the preceding layer and we ignore them.
	
	\subsection{Channel scaling through inversely-proportional factorization} \label{sec:Channels Re-scaling}
	Consider a convolutional layer with kernel W, bias B, input X, and output Y. For notional simplicity we eschew convolutions and consider matrix multiplication. To this end we denote $X^{i,j}_{1\times(K_x\cdot K_y\cdot F_{in})},Y^{i,j}_{1\times F_{out}}$, the channel vectors when the kernel is centered on spatial position i,j within X,Y. We can then write the kernel as a matrix $W_{(K_x\cdot K_y\cdot F_{in})\times F_{out}}$ and the bias as a vector $B_{1\times F_{out}}$. The following two factorizations hold:
		\small \begin{equation}  \label{eq:inputScaling}
	\begin{aligned}
	Y^{i,j} &= (X^{i,j}C_1 ) (C_1^{-1}  W) + B  
	\end{aligned}
	\end{equation} \normalsize
	
			\small \begin{equation}  \label{eq:outputScaling}
	\begin{aligned}
	Y^{i,j}C_2 &= X^{i,j}  (W  C_2) + (B  C_2)  
	\end{aligned}
	\end{equation} \normalsize
	
	where for both cases $C_i$ is a diagonal matrix with positive entries and $C_i^{-1}$ is its the inverse diagonal matrix. The first factorization scales the channels of the layer's input and the second factorization scale the channels of layer's output. 
	We now consider a simple setting where $L1$ and $L2$ are two consecutive convolutionalal layers in a network.	We assume that the activation function of $L1$, $A(\cdot )$,  is homogeneous with degree 1 for positive numbers. That is, it satisfies equation \ref{eq:multiplication invariant functions}.

	\small \begin{equation} \label{eq:multiplication invariant functions}
	A(\alpha \cdot x) = \alpha \cdot A(x) \ \ \  \forall  \alpha > 0
	\end{equation} \normalsize
	
\begin{figure}[ht]
		\vskip 0.2in
		\begin{center}
			\centerline{\includegraphics[width=\columnwidth]{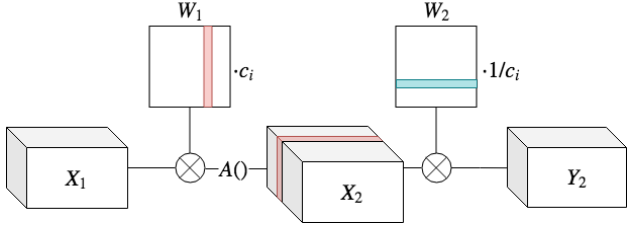}}			
			\caption{A schematic showing inversely proportional factorization in a pair of convolutional layers. An output channel $i$ in $K_1$, the kernel of $L1$, is scaled by a positive factor $c_i$. Assuming the activation function $A(\cdot)$ is homogeneous, the post activation channel is also scaled by $c_i$. All the weights in $K_2$, the kernel of $L2$ that interact with channel $i$ are inversely scaled. $Y_2$, the output of $L2$ remains unchanged. The bias terms were omitted for clarity.  }
			\label{Fig:Two layers equalization example}
		\end{center}
		\vskip -0.2in
	\end{figure}

		With this assumption, if $Y_1$ is the output of $L1$ and $X_2$ is the input to $L2$, the scaling of $Y_1$ results in a corresponding scaling of $X_2$. Combining all of the above we arrive to our main result - the post-activation output channels of $L1$ can be scaled by any positive factors by scaling the weights in the kernel and bias of $L1$ (\ref{eq:inputScaling}) and the output of the network will remain unchanged if we inversely scale the corresponding weight in $L2$ (\ref{eq:outputScaling}). We term this endomorphism an inversely-proportional factorization and it is shown schematically in Figure \ref{Fig:Two layers equalization example}.
	
	With the exception of the last layer, we can scale the individual channels withing each layer in a full network by iteratively factorizing pairs of layers. Finally, we note that the commonly used ReLU\cite{Nair:2010:RLU:3104322.3104425}, PRelu\cite{He2015} and linear activations all satisfy the homogeneous property from equation (\ref{eq:multiplication invariant functions}). Thus our scheme is applicable to most commonly used CNNs. 
	
	\subsection{Quantization Noise Analysis} 
\begin{figure}[ht]
		\vskip -0.0in
		\begin{center}
			\centerline{\includegraphics[width=\columnwidth]{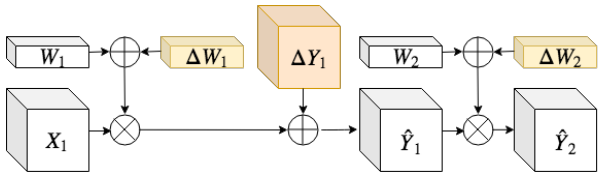}}				
			\caption{A simple model showing the effects of quantization on two layers model. $\Delta W_1$,  $\Delta W_2$, and  $\Delta Y_1$ emulate the quantization effect.}
			\label{Fig:Two layers noise model} 
		\end{center}
		\vskip -0.2in
	\end{figure}

	Understanding the formation and propagation of quantization noise across the network is an essential step in the design of better quantization algorithms. In this section we analyze the effects of weight and activations quantization using the same two-layers setting depicted in Figure \ref{Fig:Two layers equalization example}. We model the effect of limited precision by adding noise terms $\Delta W_1\in {\rm I\!R}^{K_1\times K_1\times F_{in1}\times F_{out1}}$, $\Delta W_2\in {\rm I\!R}^{K_2\times K_2\times F_{in2}\times F_{out2}}$, and $\Delta Y_1\in {\rm I\!R}^{F_{out1}}$ to $W_1$, $W_2$, and $Y_1$ respectively. The noisy model is shown in Figure \ref{Fig:Two layers noise model}. For simplicity we also assume that all the biases are zero and that all activations are linear.

	We now show how each noise term affects the overall noise at the output of each layer. For a given tensor $T$ we denote  $\hat{T}$ the noisy version of that tensor. In addition, we denote with $\tilde{T}^{\Delta N}$ the additive noise source to $T$ due to $\Delta N$. We start by calculating the output of the first layer  
	
	\vspace{-0ex}
	\begin{equation} \label{eq:Y1Calculation}
	\begin{split}
		&\hat{Y_1} = X_1\circledast (W_1 + \Delta W_1) +\Delta Y_1 \\
		&= X_1*W_1 + X_1\circledast \Delta W_1 + \Delta Y_1\\
		& \equiv Y_1 + \tilde{Y_1}^{\Delta W_1} + \tilde{Y_1}^{\Delta Y_1}
	\end{split}
	\end{equation}
	 
	The output of the first layer has two noise sources. $\tilde{Y_1}^{\Delta W_1}$ is due to the interaction of the weight quantization noise with the input $X_1$ and $\tilde{Y_1}^{\Delta Y_1}$ is the intrinsic quantization  noise. Assuming the noises are independent, their variance is:
	
	\begin{equation} \label{eq:Y1NoiseVarEstimation}
	\begin{split} 
		E \left \{ (\tilde{Y}_1^{\Delta W_1})^2 \right \} &= K_1^2\cdot F_{in1}\cdot E\{(X_1)^2\}\cdot  E\{\Delta (W_1)^2\} \\
		E \left \{ (\tilde{Y_1}^{\Delta Y_1})^2 \right \} &= E\{\Delta Y_1^2\}
	\end{split}
	\end{equation}
	
	For uniform quantization, the noise terms $\Delta W1, \Delta Y1$ distribution can be approximated as uniform, zero-centered, i.i.d processes \cite{Lin2015,Marco2005}. Denoting by $W_1^r$, $Y_1^r$ the dynamic ranges of $W_1$, $Y_1$ we get
	
	\begin{equation} \label{eq:distributionY1Y2}
	\begin{aligned} 
		&\Delta W1 \sim U(-\frac{W_1^r}{2^{N+1}},\frac{W_1^r}{2^{N+1}}) \\
		&\Delta Y1 \sim U(-\frac{Y_1^r}{2^{N+1}},\frac{Y_1^r}{2^{N+1}}) \\
	\end{aligned}
	\end{equation} 
	 	
	The dynamic range of a tensor is determined by the extreme values across all the channels within it and so the noise distribution is determined by, at most,  two channels - the channel with the largest value and the channel with smallest value. We term these channels the dominant channels and note that there is substantial variance between the extermum values of different channels.
	Next, we calculate the output of the second layer. Explicit calculation of $Y_2$ shows four noise sources, three are due to the quantization of $\Delta W_1$, $\Delta W_2$, $\Delta Y_1$, and one rooted in the multiplication of $Y_1^{\Delta W_1}$,  $Y_1^{\Delta Y_1}$ by $\Delta W_2$. The last component can be neglected in most practical scenarios and the variances of the others are
	        
	\begin{equation} \label{eq:Y2NoiseVarEstimation}
	\begin{split} 
	E \left \{ (\tilde{Y_2}^{\Delta W_1})^2 \right \} &= \left (\sum W_2^2 \right ) \cdot E \left \{ (\tilde{Y_1}^{\Delta W_1})^2 \right \} \\
	E \left \{ (\tilde{Y_2}^{\Delta Y_1})^2 \right \} &= \left (\sum W_2^2 \right ) \cdot E \left \{ (\tilde{Y_1}^{\Delta Y_2})^2 \right \} \\
	E \left \{ (\tilde{Y_2}^{\Delta W_2})^2 \right \} &= K_2^2\cdot F_{in2}\cdot E\{(Y_1)^2\}\cdot  E\{(\Delta W_2)^2\} \\
	\end{split}
	\end{equation}
	
	\subsection{Effect of inversely-proportional factorization on SQNR}\label{sec:SQNR scaling}
We use SQNR to quantify the effect of the quantization noise. 
	\begin{equation} \label{eq:sqnrDef}
		SQNR_Y \equiv \frac{E \left \{Y^2 \right \}}{E \left \{ (Y-\hat{Y})^2 \right \}} = \frac{E \left \{Y^2 \right \}}{E \left \{ \tilde{Y}^2 \right \}}  
	\end{equation}
	
	We calculate $SQNR_{Y1}$ and $SQNR_{Y2}$, the SQNRs at the outputs of layer 1 and 2 respectively, by plugging (\ref{eq:Y1NoiseVarEstimation}),(\ref{eq:Y2NoiseVarEstimation}) into (\ref{eq:sqnrDef}).

	\begin{equation} \label{eq:sqnrY1Y2}
		\begin{aligned} 
			&SQNR_{Y1} = \frac{E \left \{ Y_1^2 \right \}}{E \left \{ (\tilde{Y_1}^{\Delta W_1}+\tilde{Y_1}^{\Delta Y_1})^2 \right \}} \\
			&SQNR_{Y2} = \frac{E \left \{ Y_1^2 \right \}}{E \left \{ (\tilde{Y_2}^{\Delta W_1}+\tilde{Y_2}^{\Delta Y_1}+\tilde{Y_2}^{\Delta W_2})^2 \right \}}
		\end{aligned}
	\end{equation}
	
	We now show how inversely proportional factorization affects the SQNR of both layers. We start by looking at the effect on the signal components. We denote the scaling vector with $C\in {\rm I\!R}^{F_{out1}}$ and the scaled version of tensor $T$ by $T'$. For simplicity, since scaling the whole layer by a constant has no effect on the SQNR, we can assume without loss of generality that $C\geq 1$. Therefore, we can say that all channels in $Y_1$ are either amplified or unchanged. In addition, we showed in Section \ref{sec:Channels Re-scaling} that the factorization has no effect on $Y_2$. Thus for the signal components we have
	\begin{equation} \label{eq:equal_sig_inquality}
	\begin{aligned} 
	E\{(Y_1')^2\} &\ge E\{(Y_1)^2\} \\
	E\{(Y_2')^2\} &= E\{(Y_2)^2\}	\\
	\end{aligned}
	\end{equation}
	
	Amplifying the channels of $Y_1$ haphazardly might increase the layer's extremum values which will increases the variance of the noise sources $\Delta W_1$, $\Delta Y_1$ (\ref{eq:distributionY1Y2}). On the other hand, amplification $Y_1$ is compensated by attenuation of $W_2$ which may only decrease the variance of $\Delta W_2$ if it results in a reduction of $W_2$'s extremum values. Thus for the noise sources we have
	
	\begin{equation} \label{eq:equal_noise_inquality}
	\begin{aligned} 
		E\{(\Delta W_1')^2\} &\geq E\{(\Delta W_1)^2\} \\
		E\{(\Delta Y_1')^2\} &\geq E\{(\Delta Y_1)^2\} \\
		E\{(\Delta W_2')^2\} &\leq E\{(\Delta W_2)^2\} \\
	\end{aligned}
	\end{equation}
	
	 We now make the crucial assumption that the channels of $Y_1$ are amplified in such a manner that the variances of $\Delta W_1$, $\Delta Y_1$ remain unchanged while the variance of $\Delta W_2$ decreases. 
	 Under these assumptions, $\tilde{Y_1}^{\Delta W_1}$, $\tilde{Y_1}^{\Delta Y_1}$ are unaffected by the amplification of $Y_1$ and we get that 
	 	\begin{equation}
	 \begin{aligned} 
	 SQNR'_{Y1} \geq SQNR_{Y1}		
	 \end{aligned}
	 \end{equation}
	 
	 The effect on $SQNR_{Y2}$ (\ref{eq:sqnrY1Y2}) is more tricky. $E \left \{ (\tilde{Y2}^{\Delta W1})^2 \right \}$, $E \left \{ (\tilde{Y2}^{\Delta Y1})^2 \right \}$ are decreased by the attenuation of $W_2$. What happens to $E \left \{ (\tilde{Y_2}^{\Delta W_2})^2 \right \}$ depends on $E\{(Y_1)^2\}\cdot  E\{(\Delta W_2)^2\}$, i.e. whether the amplification of $Y_1$ is more dominant than the attenuation of $\Delta W_2$. Thus there is no guarantee that $SQNR_{Y2}$ improves.
	 Undaunted, in the next section we present a greedy algorithm that through iterative application of inversely-proportional factorization improves the SQNR across the network and reduces the post-quantization degradation.

	\section{Equalization Algorithm} \label{sec:Equalization Algorithm}
\begin{figure*}
	\centering     
	\subfigure[One Step Equalization]{\label{fig:a}\includegraphics[width=\columnwidth]{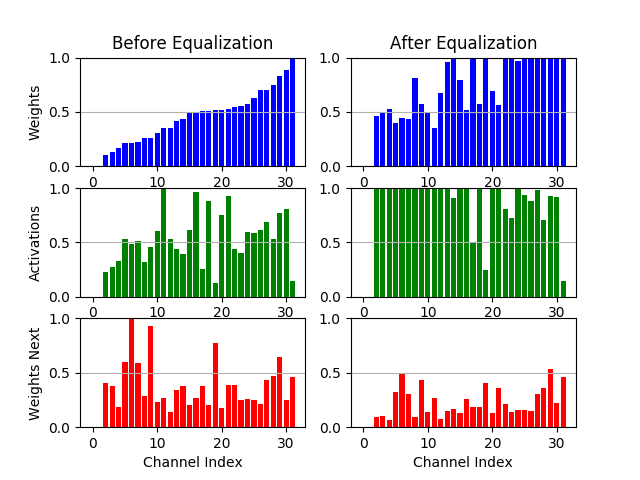}}
	\subfigure[Two Steps Equalization]{\label{fig:b}\includegraphics[width=\columnwidth]{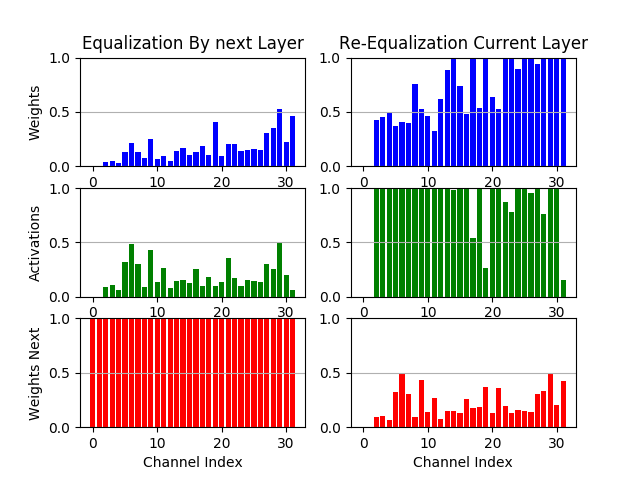}}	
	\caption{Example of the effect of channel equalization on the two layers model. Each bar indicates the maximum values of a channel. The blue graphs show the data for weights, the green for the activations, and the red the weights of the successor layer. Figure (a) demonstrates the one step algorithm. On the left is the initial state and on the right is the state after equalization. Figure (b) demonstrates the two steps algorithm. On the left is the state after is the next layer weights equalization and on the right is the final state}
	\label{Fig:EqualExample}
	\end{figure*}

	Building on the analysis in Section \ref{sec:Theoretical Foundation} we propose two algorithms designed around the idea of applying a pre-quantization factorization that increases the energy of channels without changing the variance of the noise. This is achieved by amplifying non-dominant channels such that their extremum values are matched with those of the dominant channels. As shown in Figure \ref{Fig:EqualExample}, these algorithms tends to equalize the channels' energy and therefore got the name \textit{channel equalization}. 
	
	\subsection{One-Step Equalization} \label{sec:One Step Equalization}
		\begin{algorithm}
		\caption{One Step Equalization}\label{alg:One Step Equalization}
		\begin{algorithmic}[1]			
			\WHILE {1}
			\STATE $layer \gets getNextLayer()$
			\IF {layer is last layer}
			\STATE return
			\ENDIF
			\STATE $kerOutChMax \gets getKerOutChMax(layer)$
			\STATE $actOutChMax \gets getActOutChMax(layer)$
			\STATE $kerOutMax \gets max(kerOutChMax)$
			\STATE $actOutMax \gets max(actOutChlMax)$
			\STATE $kerScale \gets \frac{kerOutMax}{kerOutChMax}$
			\STATE $actScale \gets \frac{actOutMax}{actOutChMax}$
			\STATE $scale \gets min(kerScale,actScale,S_{max})$
			\STATE $scaleLayer(layer, scale)			$
			\ENDWHILE
		\end{algorithmic}	
	\end{algorithm} 
   
	Algorithm \ref{alg:One Step Equalization} explains a simple, one-step channel equalization method. We assume that the network can be represented as direct acyclic graph with layers being represnted by nodes and that it is topologicaly sorted. The algorithm is then applied iteratively beginning at the first layer(node) and continues until we reach all of the network's output layers. At each iteration the layer's channels are equalized by employing inversely-proportional factorization with its successor layers. A layer is eligible to be equalized only once all of its predecessor layers were equalized. The function $getNextLayer()$ returns the next layer. The functions $getKerOutChMax(layer)$ and $getActOutChMax(layer)$ return the maximum values per channel for the weights and activation respectively. Each one of them results in a vector of length $F_{out}$. For each channel, we calculate the ratio between the layer's extermum and the channel's extremum for the activations and weights, these ratios are defined as the activation and weight scales. When a layer is scaled, each channel is scaled by the minimum between the activation and weight scales. We further limit the scale by a pre-defined maximum to prevent the over-scaling of channels with small activation values.  $scaleLayer(layer, scale)$ is the scaling of the layer and its successors according to (\ref{eq:inputScaling}), (\ref{eq:outputScaling}). It is easy to see that the maximum values of the weights and the activation post equalization won't change and that all the scales are $\geq1$. 
	
	An example of the results of one-step equalization on the channel scales within a layer is shown in Figure \ref{Fig:EqualExample}(a). We see that, post equalization, channels have much less variance in scales which in turn implies that they tend to have similar energy.
    As explained in section \ref{sec:SQNR scaling} for each iteration there is no guarantee that $SQNR_{Y2}$ improves but in most cases we witnessed that it did. Moreover, even if the $SQNR_{Y2}$ decreases it will improved in next iteration when the channel of $Y_2$ will be equalized.

	\subsection{Two Steps Equalization} \label{sec:Two Steps Equalization}
	We define optimal equalization (OE) as the state where the extremum values of all the channels are equal. OE can be done in terms of weights only, activations only or both. OE for activations or weights can always be achieved but equalization of one will be sub-optimal for the other. OE for both, on the other hand, is out of reach in most cases because we have only one scale per channel. The two steps equalization tries to make a step toward the optimal OE.	
	
		\begin{algorithm}
		\caption{Two Steps Equalization}\label{alg:Two Steps Equalization}
		\begin{algorithmic}[1]			
			\WHILE {1}
			\STATE $layer \gets getNextLayer()$
			\IF {layer is last layer}
			\STATE return
			\ENDIF
			\STATE $sucInChMax \gets getSucInChMax(layer)$
			\STATE $sucInMax \gets max(sucInChMax)$
			\STATE $kerOutChMax \gets getKerOutChMax(layer)$
			\STATE $actOutChMax \gets getActOutChMax(layer)$
			\STATE $kerOutMax \gets max(kerOutChMax)$
			\STATE $actOutMax \gets max(actOutChMax)$
			\STATE $kerScale \gets \frac{KerOutMax}{kerOutChMax}\cdot \frac{sucInChMax}{sucInMax}$
			\STATE $actScale \gets \frac{actOutMax}{actOutChMax}\cdot \frac{sucInChMax}{sucInMax}$
			\STATE $scale = min(kerScale,actScale,S_{max})$
			\STATE $scale = scale/min(scale)$
			\STATE $scaleLayer(layer, scale)			$
			\ENDWHILE
		\end{algorithmic}	
	\end{algorithm}

	Algorithm \ref{alg:Two Steps Equalization} explains the two steps equalization process. The basic idea is to diminish the layers extremum values before the equalization. This is done by applying proportionally-inverse factorization in reverse - we attenuate the channels of the first layer and compensate by amplifying the values of the second layer. To avoid increasing the weight noise in the second layer the compensating amplification is not allowed to change the extremum values of the second layer. This is done by using the function $getSucInChMax$ that returns the maximum per input channel of the successor layer weights. Dividing each channel by $\frac{sucInChMax}{sucInMax}$ will equalize the next layer and attenuate all the channels in the current layer. The second step of the algorithm is the same as in the one-step equalization. At the end of the algorithm we normalized all the scales so they will all be $\geq 1$. Figure \ref{Fig:EqualExample}(b) shows an example of two steps equalization. We can see that the channels are equalized a little bit better and that the maximum value of the next layer is lower. Therefore, we can expect the noise in the network after this equalization will be attenuated and indeed our tests showed that this algorithm can produce better channel equalization. Our intuition is that if a dominant channel can be attenuated than it means that the weights of second layer multiplying it are small. In other words - before the factorization the second layer was "naturally" attenuating the channel, signaling that its scale is too large compared to the other channels. In a limited precision setting it is important that this "gain control" be done beforehand since quantization is adversely affected by channels with outlier scales.

	\section{Experiments and Results} \label{sec:Results}
	In this section we perform experiments analyzing the performance of our proposed algorithms. We first verify our analysis in previous sections by measuring the noise across test networks with and without equalization. We then show that a reduction in noise translates to a reduction in the post-quantization degradation of classifier networks trained on the ImageNet dataset\cite{ILSVRC15} and finally we show that our algorithm can also be applied to MobileNets\cite{Howard2017} with some modifications. In all our tests we used layer wise quantization. Activations were encoded using 8-bit unsigned integers and weights were encoded using 8-bit integers. Biases were encoded using 16-bits integers. We used passive quantization, meaning that no retrain was used and there is no need for labeled data. For all experiments we extract the activation extremum values using 64 images.
	
	\subsection{SQNR measurements}
	We designed a test that shows the noise of each layer separately. Moreover, the test can differentiate between activations and weights noises. For each layer we measured three quantities: the layer output where only the weights are quantized ($Y_1 + \tilde{Y_1}^{\Delta W_1}$), the layer output where only the activations are quantized($Y_1 + \tilde{Y_1}^{\Delta Y1}$), and the layer output where both are quantized ($\hat{Y_1}$). To measure the noise we compared these quantities to those of the  original full precision layer output. 
	
\begin{figure}[h]
		\centering
		\includegraphics[width=\columnwidth]{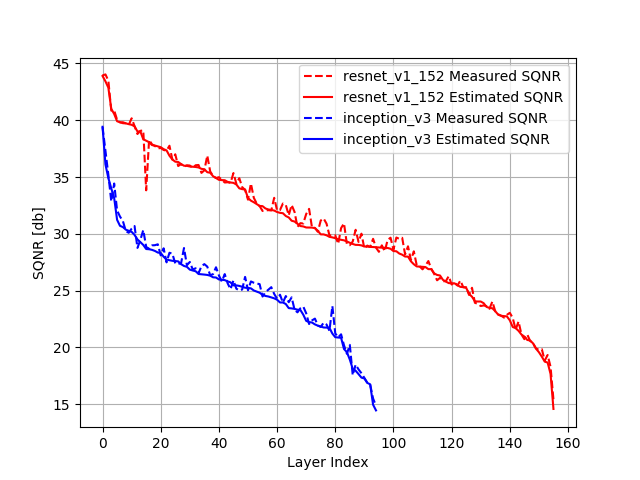}			
		\caption{Noise estimation verification, A Comparison between the estimated SQNR according to (\ref{eq:Y1NoiseVarEstimation}) and measured SQNR. The layer index is sorted by the estimated SQNR from largest to smallest.}
		\label{Fig:estimatedNoises}
	\end{figure}

	We measured the SQNR of each layer and compared it to the one predicted by (\ref{eq:Y1NoiseVarEstimation}) to verify our assumptions. The results of this experiment are shown in in Figure \ref{Fig:estimatedNoises} for ResNet-152\cite{He2015b} and Inception-V3\cite{Szegedy2015}. The results show good agreement between predicted and measured noise. 
	
\begin{figure}[h]
		\centering
		\includegraphics[width=\columnwidth]{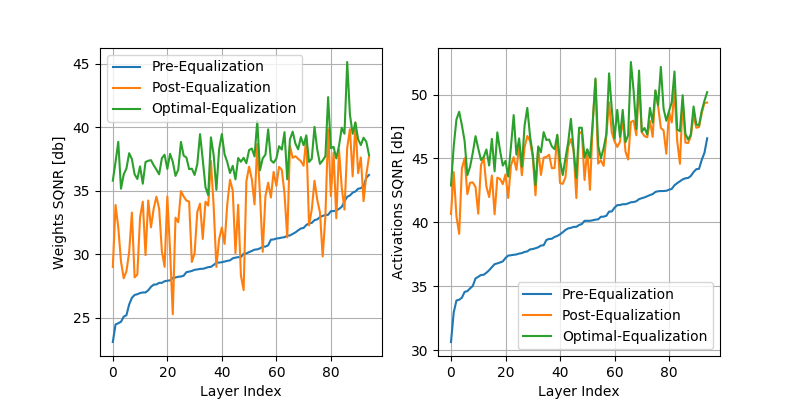}		
		\caption{A comparison to the optimal SQNR. Demonstrating the effect of equalization algorithm on the SQNR for inception-V3. The figure shows how the algorithm tries to close the gap between the original state and optimal equalization. In these graphs the energy of the signal is measured and noise is estimated based on a uniform distribution assumption. The layer index is sorted by the estimated SQNR from largest to smallest.}
		\label{Fig:OptEqualComapision}
	\end{figure}

	We now analyze the quality of our two-steps equalization algorithm. To that end we look at a simple setting where only the weight or only the activations are quantized. For weight quantization, we compare $SQNR_{W1}\equiv \sum (W_1)^2 / \sum (\Delta W_1)^2$, a measure of how well the layers weights are equalized to the weight OE. And for activation quantization, we compare $SQNR_{Y1}\equiv \sum (Y_1)^2 / \sum (\Delta Y_1)^2$ , a measure of how well the layers output channels are equalized, to the activation OE. This gives us an idea how far we are from the overall OE of both activations and weights. Figure \ref{Fig:OptEqualComapision} shows the results of this comparison on Inception-V3. We see that the method improves significantly the SQNR of the activations and almost reached the performance of the OE. For weights, the effect is smaller and the gap to the OE is larger.

\begin{figure}
	\centering     
	\subfigure[Matched]{\label{fig:Inception-V3}\includegraphics[width=0.45\columnwidth]{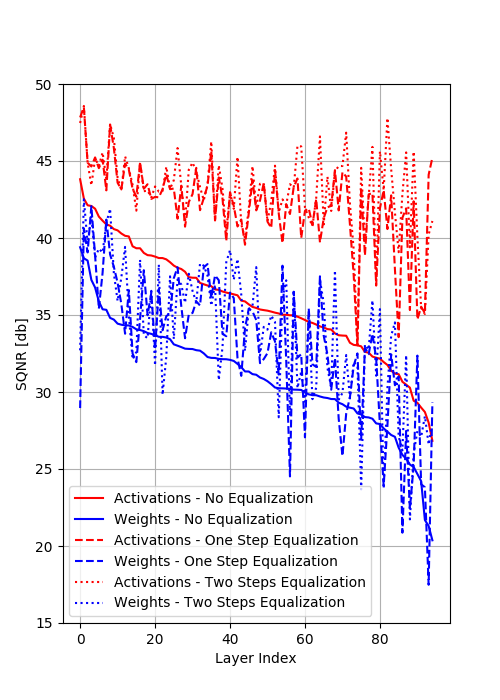}}
	\subfigure[Sorted]{\label{fig:Inception-V3 Sorted}\includegraphics[width=0.45\columnwidth]{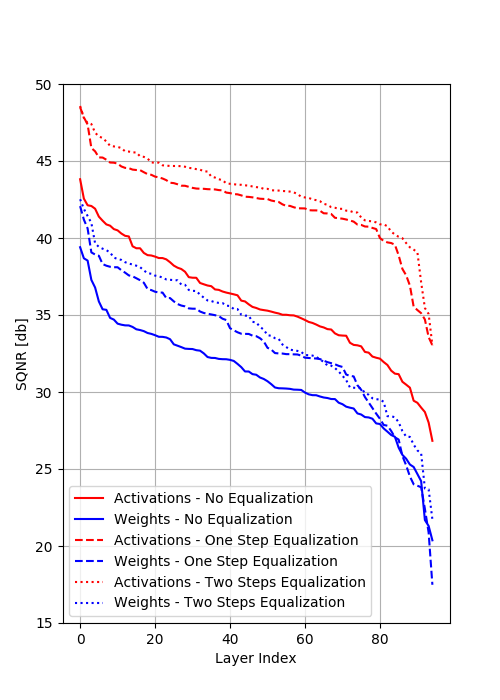}}
	\caption{Noise measurement across the network for Inception-V3. Measure the activation SQNR $\sum (Y_1)^2/\sum (\tilde{Y}_1^{\Delta Y_1})^2$ and weights SQNR $\sum (Y_1)^2/\sum (\tilde{Y}_1^{\Delta W_1})^2$ for three cases: pre-equalization, post one step equalization, and post two steps equalization. In figure (a), for better visibility, the layers are sorted by pre-equalization SQNR values. In figure (b), for even better visibility, the post equalization lines are sorted as well.}
	\label{Fig:equalizationComperasion}
\end{figure}

	Finally, we measured the effect of equalization on noises throughout the network. We measured the SQNR in three cases: without equalization, with one- and two-steps equalization. Observing the results, as shown on Figure \ref{Fig:equalizationComperasion}, several conclusions can be drawn:(1) overall the SQNR is improved by equalization (2) The greedy nature of the algorithm means that for some intermediate layers the weight induced SQNR decreases. This is due to the fact that weight induced noise is increased when the layer input is amplified.

	\subsection{ImageNet Quantization Performance}
	We now show that the overall reduction in SQNR achieved by the equalization algorithms results in improved performance of the quantized networks. Table \ref{tbl:QuantizationDegradationComparison} summarizes our results. We compared the classification performances of the quantized networks to their floating point version. Equalization improves quantization performance for almost all nets and for some, like inception-V1 \cite{Inception_v1} and DenseNet121 , a considerable improvement is observed. An examination of these networks revealed layers which suffer from an extreme imbalance between the channels. As we showed, this significantly increase the noise within the layer, triggering an avalanche effect throughout the network. Equalization eliminates this effect and enables good performance post quantization. Overall we see that two-steps equalization gives better performance than one-step equalization.

	Figure \ref{Fig:equalizationComperasion} shows that the noise stemming from activation quantization is always reduced while the noise stemming from weight quantization occasionally increases after equalization. To quantify the effect we have on both noise sources we repeated the performance measurements for weights-only and activations-only quantization. We see that equalization has a positive effect for both scenarios and that which noise term is dominant is network dependent. For example, for the Inception architectures, weights quantization is dominant, while for ResNet-152 and DenseNet activation quantization is dominant. In addition, our results indicate that the total degradation $\approx$ weight degradation + activation degradation. 

	MobileNet is a challenging architecture for quantization and many passive quantization schemes result in large degradation \cite{Jacob2017,Lee2018}. It also employs ReLU6\cite{Jacob2017} activations which require special treatment for the equalization algorithm to work since it does not satisfy (\ref{eq:multiplication invariant functions}). 
			\begin{table}[]
					\caption{A comparison of the performance of the two proposed algorithms. For each network we tabulate the post-quantization degradation without equalization and with one- and two-steps equalization. The degradation is measured relative to the networks top-1 accuracy which is estimated using the full ILSCRC2015 validation set of 50K images. For Resnet-152 and Inception-V3 we also measure the degradation when only the weights or activations are quantized. To facilitate reproducibility we used open source pre-trained networks. All networks are taken from TF-slim models \cite{TFSilm} with the exception of DenseNet\cite{DenseNetGit}  }
		\begin{tabular}{|l||l||l|l|l|}
		\hline
		\textbf{\begin{tabular}[c]{@{}l@{}}Network \\ Name\end{tabular}} &
		\textbf{\begin{tabular}[c]{@{}l@{}} Original  \\ Top-1 \\ accuracy \end{tabular}} &
		\textbf{\begin{tabular}[c]{@{}l@{}}No \\ Equal\end{tabular}} & \textbf{\begin{tabular}[c]{@{}l@{}}One \\ Step \\ Equal\end{tabular}} & \textbf{\begin{tabular}[c]{@{}l@{}}Two \\ Steps\\  Equal\end{tabular}} \\ \hline
		\multicolumn{5}{|c|}{Weights and Activations Quantization} \\ \hline
		ResNet-V1-50  &$75.2\%$& $0.25\%$   & $0.38\%$ & $0.25\%$ \\ \hline
		ResNet-V1-152 &$76.8\%$& $1.27\%$   & $0.80\%$ & $0.78\%$ \\ \hline
		Inception-V3  &$77.9\%$& $0.66\%$   & $0.47\%$ & $0.35\%$ \\ \hline
		Inception-V1  &$69.8\%$& $2.23\%$   & $0.41\%$ & $0.39\%$ \\ \hline
		DenseNet-121  &$74.3\%$& $5.78\%$   & $0.42\%$ & $0.35\%$ \\ \hline
		\multicolumn{5}{|c|}{Weights Quantization Only} \\ \hline
		ResNet-V1-152 &$76.8\%$& $0.23\%$   & $0.19\%$ & $0.16\%$ \\ \hline
		Inception-V3  &$77.9\%$& $0.55\%$   & $0.36\%$ & $0.32\%$ \\ \hline
		\multicolumn{5}{|c|}{Activations Quantization Only} \\ \hline
		ResNet-V1-152 &$76.8\%$& $0.98\%$   & $0.66\%$ & $0.62\%$ \\ \hline
		Inception-V3  &$77.9\%$& $0.09\%$   & $0.03\%$ & $0.05\%$ \\ \hline
		
	\end{tabular}
	
	\label{tbl:QuantizationDegradationComparison}
\end{table}		 
	One-step equalization can be used almost without change if it only amplifies channels with extremum activation value below 6. One-step equalization gives limited improvement (Table \ref{tbl:QuantizationDegradationComparisonMobileNet}). To enable Two-steps equalization we disable the division by $min(scale)$ at the end of algorithm \ref{alg:Two Steps Equalization} and in doing so allow the algorithm to attenuate channels ($C<1$). However, to prevent significant modification to the full-precision network results, we limit the attenuation to 70\% of the original range. This method shows negligible effect on the full-precision network but shows a significant improvement for the quantized network (Table \ref{tbl:QuantizationDegradationComparisonMobileNet}).

	 In addition, we found that due to the use of depthwise convolutions which have a small number of weights in each kernel the mean of the quantized weights might be different from the original value which results in a shift of the distribution. As a remedy to this problem we use knowledge distillation \cite{Hinton2015} and fine-tune only the biases to compensate for the shift so that the distribution means are the same for both the original and quantized network. Since only the biases are being updated 1000 unlabeled images are all that is needed and the fine-tuning process is very short. Used in conjunction with equalization we get competitive results (Table \ref{tbl:QuantizationDegradationComparisonMobileNet}) with the state-of-the art \cite{Jacob2017,Lee2018,Krishnamoorthi2018,Sheng2018,Rub1}. However, our result is unique in the following regards: it doesn't require channel-wise quantization which has significant overhead for hardware implementation as well as additional storage requirements. It uses only $\sim 1000$ unlabeled images allowing it to be used with off-the-shelf pre-trained models and the quantization process is simple and very fast to implement.
	
	\begin{table}[]
		\caption{MobileNet degradation of the different quantization schemes compared to the floating point implementation. All networks are taken from TF-slim models \cite{TFSilm}. }
		\begin{tabular}{|l|l|l|l|l|l|}
			\hline
			\textbf{\begin{tabular}[c]{@{}l@{}}Net \\ Version\end{tabular}} & \textbf{\begin{tabular}[c]{@{}l@{}}No \\ Equal\end{tabular}} & \textbf{\begin{tabular}[c]{@{}l@{}}One \\ Step \\ Equal\end{tabular}} & \textbf{\begin{tabular}[c]{@{}l@{}}Two \\ Steps\\  Equal\end{tabular}} &
			\textbf{\begin{tabular}[c]{@{}l@{}}Bias \\  Only\end{tabular}} & \textbf{\begin{tabular}[c]{@{}l@{}} Equal+\\ Bias \end{tabular}} \\ \hline
			\begin{tabular}[c]{@{}l@{}}V1-1.0\end{tabular}       &$7.89\%$  & $6.12\%$  & $3.2\%$ & $1.3\%$ & $0.95\%$\\ \hline
			\begin{tabular}[c]{@{}l@{}}V2-1.0\end{tabular}       &$42.68\%$  & $4.07\%$ & $2.1\%$ & $1.5\%$ & $0.61\%$\\ \hline
			\begin{tabular}[c]{@{}l@{}}V2-1.4\end{tabular}       & $8.06\%$ & $6.21\%$  & $1.9\%$ & $1.4\%$ & $0.55\%$ \\ \hline
		\end{tabular}
		\label{tbl:QuantizationDegradationComparisonMobileNet}
	\end{table}

\section{Discussion}

This paper highlights a property of convolutional neural networks that is often overlooked which allows inversely-proportional factorization. We showed methods to harness this property to generate equivalent networks that are much more robust to quantization noises. Our intuition was that networks have implicit "gain control" mechanisms that can be made explicit through channel equalization. When the channel are equalized, outliers are removed and quantization performance is improved. Given the same constrains of 8bits quantization, layer-wise scaling, and without re-training our algorithms reached state-of-the-art performance. 

Our focus was on passive quantization allowing rapid deployment, however, equalization should benefit other quantization methods. When fine tuning or quantization-aware training are used, equalization can be integrated as a pre-processing step to reduce noise prior to training. We believe that most current quantization methods will benefit from applying proper equalization.  

There is much more to explore towards realizing the full potential of inversely proportional factorizations. We suggested a greedy equalization algorithm that performs well but advanced equalization algorithms can push the improvement even further. For example, we showed that the impact of weights and activations quantization might change between layers. This property can be exploited for better equalization. In addition, advanced prediction methods of the noise's effect on the network performance like those suggested in \citet{pmlr-v70-sakr17a} or \citet{DBLP:journals/corr/ChoiEL16} can be used for equalization optimization.  
  
More generally, this work is a first attempt to utilize equivalent net factorizations. The approach should find merit in other applications as well. For pruning, activation only equalization can be employed to make the interpretation of weight importance more natural. After equalization, small weights have less effect on the network and therefore are more likely to be pruned by methods that rely on the relative weight size\cite{Han2015}. During training, inversely-proportional factorization can be used to scale the gradients of different channels within a layer allowing for faster convergence or avoiding vanishing/exploding gradients.

\bibliography{equalizationMain}

\begin{thebibliography}{35}
\providecommand{\natexlab}[1]{#1}
\providecommand{\url}[1]{\texttt{#1}}
\expandafter\ifx\csname urlstyle\endcsname\relax
  \providecommand{\doi}[1]{doi: #1}\else
  \providecommand{\doi}{doi: \begingroup \urlstyle{rm}\Url}\fi

\bibitem[Banner et~al.(2018{\natexlab{a}})Banner, Hubara, Hoffer, and
  Soudry]{Banner2018a}
Banner, R., Hubara, I., Hoffer, E., and Soudry, D.
\newblock {Scalable Methods for 8-bit Training of Neural Networks}.
\newblock may 2018{\natexlab{a}}.
\newblock URL \url{http://arxiv.org/abs/1805.11046}.

\bibitem[Banner et~al.(2018{\natexlab{b}})Banner, Nahshan, Hoffer, and
  Soudry]{Banner2018b}
Banner, R., Nahshan, Y., Hoffer, E., and Soudry, D.
\newblock {ACIQ: Analytical Clipping for Integer Quantization of neural
  networks}.
\newblock oct 2018{\natexlab{b}}.
\newblock URL \url{http://arxiv.org/abs/1810.05723}.

\bibitem[Choi et~al.(2018)Choi, Wang, Venkataramani, Chuang, Srinivasan, and
  Gopalakrishnan]{Choi2018}
Choi, J., Wang, Z., Venkataramani, S., Chuang, P. I.-J., Srinivasan, V., and
  Gopalakrishnan, K.
\newblock {PACT: Parameterized Clipping Activation for Quantized Neural
  Networks}.
\newblock may 2018.
\newblock URL \url{http://arxiv.org/abs/1805.06085}.

\bibitem[Choi et~al.(2016)Choi, El{-}Khamy, and
  Lee]{DBLP:journals/corr/ChoiEL16}
Choi, Y., El{-}Khamy, M., and Lee, J.
\newblock Towards the limit of network quantization.
\newblock \emph{CoRR}, abs/1612.01543, 2016.
\newblock URL \url{http://arxiv.org/abs/1612.01543}.

\bibitem[Courbariaux et~al.(2014)Courbariaux, Bengio, and
  David]{Courbariaux2014}
Courbariaux, M., Bengio, Y., and David, J.-P.
\newblock {Training deep neural networks with low precision multiplications}.
\newblock dec 2014.
\newblock URL \url{http://arxiv.org/abs/1412.7024}.

\bibitem[GitHub()]{DenseNetGit}
GitHub, I.
\newblock pudae/tensorflow-densenet.
\newblock \url{https://github.com/pudae/tensorflow-densenet}.

\bibitem[Google()]{Rub1}
Google.
\newblock Tensorflow lite.
\newblock URL
  \url{https://www.tensorflow.org/lite/performance/model_optimization}.

\bibitem[Gupta et~al.(2015)Gupta, Agrawal, Gopalakrishnan, and
  Narayanan]{Gupta2015}
Gupta, S., Agrawal, A., Gopalakrishnan, K., and Narayanan, P.
\newblock {Deep Learning with Limited Numerical Precision}.
\newblock feb 2015.
\newblock URL \url{https://arxiv.org/abs/1502.02551}.

\bibitem[Han et~al.(2015)Han, Mao, and Dally]{Han2015}
Han, S., Mao, H., and Dally, W.~J.
\newblock {Deep Compression: Compressing Deep Neural Networks with Pruning,
  Trained Quantization and Huffman Coding}.
\newblock oct 2015.
\newblock URL \url{http://arxiv.org/abs/1510.00149}.

\bibitem[He et~al.(2015{\natexlab{a}})He, Zhang, Ren, and Sun]{He2015}
He, K., Zhang, X., Ren, S., and Sun, J.
\newblock {Delving Deep into Rectifiers: Surpassing Human-Level Performance on
  ImageNet Classification}.
\newblock feb 2015{\natexlab{a}}.
\newblock URL \url{http://arxiv.org/abs/1502.01852}.

\bibitem[He et~al.(2015{\natexlab{b}})He, Zhang, Ren, and Sun]{He2015b}
He, K., Zhang, X., Ren, S., and Sun, J.
\newblock Deep residual learning for image recognition.
\newblock \emph{CoRR}, abs/1512.03385, 2015{\natexlab{b}}.
\newblock URL \url{http://arxiv.org/abs/1512.03385}.

\bibitem[Hinton et~al.(2015)Hinton, Vinyals, and Dean]{Hinton2015}
Hinton, G., Vinyals, O., and Dean, J.
\newblock {Distilling the Knowledge in a Neural Network}.
\newblock mar 2015.
\newblock URL \url{https://arxiv.org/abs/1503.02531}.

\bibitem[Howard et~al.(2017)Howard, Zhu, Chen, Kalenichenko, Wang, Weyand,
  Andreetto, and Adam]{Howard2017}
Howard, A.~G., Zhu, M., Chen, B., Kalenichenko, D., Wang, W., Weyand, T.,
  Andreetto, M., and Adam, H.
\newblock {MobileNets: Efficient Convolutional Neural Networks for Mobile
  Vision Applications}.
\newblock apr 2017.
\newblock URL \url{http://arxiv.org/abs/1704.04861}.

\bibitem[Huang et~al.(2016)Huang, Liu, van~der Maaten, and
  Weinberger]{Huang2016}
Huang, G., Liu, Z., van~der Maaten, L., and Weinberger, K.~Q.
\newblock {Densely Connected Convolutional Networks}.
\newblock aug 2016.
\newblock URL \url{http://arxiv.org/abs/1608.06993}.

\bibitem[Ignatov et~al.(2018)Ignatov, Timofte, Chou, Wang, Wu, Hartley, and
  {Van Gool}]{Ignatov2018}
Ignatov, A., Timofte, R., Chou, W., Wang, K., Wu, M., Hartley, T., and {Van
  Gool}, L.
\newblock {AI Benchmark: Running Deep Neural Networks on Android Smartphones}.
\newblock oct 2018.
\newblock URL \url{http://arxiv.org/abs/1810.01109}.

\bibitem[Ioffe \& Szegedy(2015)Ioffe and Szegedy]{IoffeS15}
Ioffe, S. and Szegedy, C.
\newblock Batch normalization: Accelerating deep network training by reducing
  internal covariate shift.
\newblock \emph{CoRR}, abs/1502.03167, 2015.
\newblock URL \url{http://arxiv.org/abs/1502.03167}.

\bibitem[Jacob et~al.(2017)Jacob, Kligys, Chen, Zhu, Tang, Howard, Adam, and
  Kalenichenko]{Jacob2017}
Jacob, B., Kligys, S., Chen, B., Zhu, M., Tang, M., Howard, A., Adam, H., and
  Kalenichenko, D.
\newblock {Quantization and Training of Neural Networks for Efficient
  Integer-Arithmetic-Only Inference}.
\newblock 2017.
\newblock \doi{10.1109/CVPR.2018.00286}.

\bibitem[Jouppi et~al.(2017)Jouppi, Young, Patil, Patterson, Agrawal, Bajwa,
  Bates, Bhatia, Boden, Borchers, Boyle, Cantin, Chao, Clark, Coriell, Daley,
  Dau, Dean, Gelb, Ghaemmaghami, Gottipati, Gulland, Hagmann, Ho, Hogberg, Hu,
  Hundt, Hurt, Ibarz, Jaffey, Jaworski, Kaplan, Khaitan, Koch, Kumar, Lacy,
  Laudon, Law, Le, Leary, Liu, Lucke, Lundin, MacKean, Maggiore, Mahony,
  Miller, Nagarajan, Narayanaswami, Ni, Nix, Norrie, Omernick, Penukonda,
  Phelps, Ross, Ross, Salek, Samadiani, Severn, Sizikov, Snelham, Souter,
  Steinberg, Swing, Tan, Thorson, Tian, Toma, Tuttle, Vasudevan, Walter, Wang,
  Wilcox, and Yoon]{Jouppi2017}
Jouppi, N.~P., Young, C., Patil, N., Patterson, D., Agrawal, G., Bajwa, R.,
  Bates, S., Bhatia, S., Boden, N., Borchers, A., Boyle, R., Cantin, P.-l.,
  Chao, C., Clark, C., Coriell, J., Daley, M., Dau, M., Dean, J., Gelb, B.,
  Ghaemmaghami, T.~V., Gottipati, R., Gulland, W., Hagmann, R., Ho, C.~R.,
  Hogberg, D., Hu, J., Hundt, R., Hurt, D., Ibarz, J., Jaffey, A., Jaworski,
  A., Kaplan, A., Khaitan, H., Koch, A., Kumar, N., Lacy, S., Laudon, J., Law,
  J., Le, D., Leary, C., Liu, Z., Lucke, K., Lundin, A., MacKean, G., Maggiore,
  A., Mahony, M., Miller, K., Nagarajan, R., Narayanaswami, R., Ni, R., Nix,
  K., Norrie, T., Omernick, M., Penukonda, N., Phelps, A., Ross, J., Ross, M.,
  Salek, A., Samadiani, E., Severn, C., Sizikov, G., Snelham, M., Souter, J.,
  Steinberg, D., Swing, A., Tan, M., Thorson, G., Tian, B., Toma, H., Tuttle,
  E., Vasudevan, V., Walter, R., Wang, W., Wilcox, E., and Yoon, D.~H.
\newblock {In-Datacenter Performance Analysis of a Tensor Processing Unit}.
\newblock apr 2017.
\newblock URL \url{http://arxiv.org/abs/1704.04760}.

\bibitem[Krishnamoorthi(2018)]{Krishnamoorthi2018}
Krishnamoorthi, R.
\newblock {Quantizing deep convolutional networks for efficient inference: A
  whitepaper}.
\newblock Technical report, 2018.

\bibitem[Lee et~al.(2018)Lee, Ha, Choi, Lee, and Lee]{Lee2018}
Lee, J.~H., Ha, S., Choi, S., Lee, W., and Lee, S.
\newblock Quantization for rapid deployment of deep neural networks.
\newblock \emph{CoRR}, abs/1810.05488, 2018.
\newblock URL \url{http://arxiv.org/abs/1810.05488}.

\bibitem[Lin et~al.(2015)Lin, Talathi, and Annapureddy]{Lin2015}
Lin, D.~D., Talathi, S.~S., and Annapureddy, V.~S.
\newblock {Fixed Point Quantization of Deep Convolutional Networks}.
\newblock nov 2015.
\newblock URL \url{https://arxiv.org/abs/1511.06393}.

\bibitem[Marco \& Neuhoff(2005)Marco and Neuhoff]{Marco2005}
Marco, D. and Neuhoff, D.~L.
\newblock The validity of the additive noise model for uniform scalar
  quantizers.
\newblock \emph{IEEE Transactions on Information Theory}, 51\penalty0
  (5):\penalty0 1739--1755, May 2005.
\newblock ISSN 0018-9448.
\newblock \doi{10.1109/TIT.2005.846397}.

\bibitem[McKinstry et~al.(2018)McKinstry, Esser, Appuswamy, Bablani, Arthur,
  Yildiz, and Modha]{McKinstry2018}
McKinstry, J.~L., Esser, S.~K., Appuswamy, R., Bablani, D., Arthur, J.~V.,
  Yildiz, I.~B., and Modha, D.~S.
\newblock {Discovering Low-Precision Networks Close to Full-Precision Networks
  for Efficient Embedded Inference}.
\newblock sep 2018.
\newblock URL \url{http://arxiv.org/abs/1809.04191}.

\bibitem[Migacz(2017)]{Migacz2017}
Migacz, S.
\newblock {8-bit Inference with TensorRT}.
\newblock 2017.
\newblock URL
  \url{http://on-demand.gputechconf.com/gtc/2017/presentation/s7310-8-bit-inference-with-tensorrt.pdf{\%}0Ahttp://on-demand.gputechconf.com/gtc/2017/video/s7310-szymon-migacz-8-bit-inference-with-tensorrt.mp4}.

\bibitem[Nair \& Hinton(2010)Nair and Hinton]{Nair:2010:RLU:3104322.3104425}
Nair, V. and Hinton, G.~E.
\newblock Rectified linear units improve restricted boltzmann machines.
\newblock In \emph{Proceedings of the 27th International Conference on
  International Conference on Machine Learning}, ICML'10, pp.\  807--814, USA,
  2010. Omnipress.
\newblock ISBN 978-1-60558-907-7.
\newblock URL \url{http://dl.acm.org/citation.cfm?id=3104322.3104425}.

\bibitem[Park et~al.(2018)Park, Naumov, Basu, Deng, Kalaiah, Khudia, Law,
  Malani, Malevich, Nadathur, Pino, Schatz, Sidorov, Sivakumar, Tulloch, Wang,
  Wu, Yuen, Diril, Dzhulgakov, Hazelwood, Jia, Jia, Qiao, Rao, Rotem, Yoo, and
  Smelyanskiy]{Park2018b}
Park, J., Naumov, M., Basu, P., Deng, S., Kalaiah, A., Khudia, D., Law, J.,
  Malani, P., Malevich, A., Nadathur, S., Pino, J., Schatz, M., Sidorov, A.,
  Sivakumar, V., Tulloch, A., Wang, X., Wu, Y., Yuen, H., Diril, U.,
  Dzhulgakov, D., Hazelwood, K., Jia, B., Jia, Y., Qiao, L., Rao, V., Rotem,
  N., Yoo, S., and Smelyanskiy, M.
\newblock {Deep Learning Inference in Facebook Data Centers: Characterization,
  Performance Optimizations and Hardware Implications}.
\newblock nov 2018.
\newblock URL \url{http://arxiv.org/abs/1811.09886}.

\bibitem[Russakovsky et~al.(2015)Russakovsky, Deng, Su, Krause, Satheesh, Ma,
  Huang, Karpathy, Khosla, Bernstein, Berg, and Fei-Fei]{ILSVRC15}
Russakovsky, O., Deng, J., Su, H., Krause, J., Satheesh, S., Ma, S., Huang, Z.,
  Karpathy, A., Khosla, A., Bernstein, M., Berg, A.~C., and Fei-Fei, L.
\newblock {ImageNet Large Scale Visual Recognition Challenge}.
\newblock \emph{International Journal of Computer Vision (IJCV)}, 115\penalty0
  (3):\penalty0 211--252, 2015.
\newblock \doi{10.1007/s11263-015-0816-y}.

\bibitem[Sakr et~al.(2017)Sakr, Kim, and Shanbhag]{pmlr-v70-sakr17a}
Sakr, C., Kim, Y., and Shanbhag, N.
\newblock Analytical guarantees on numerical precision of deep neural networks.
\newblock In Precup, D. and Teh, Y.~W. (eds.), \emph{Proceedings of the 34th
  International Conference on Machine Learning}, volume~70 of \emph{Proceedings
  of Machine Learning Research}, pp.\  3007--3016, International Convention
  Centre, Sydney, Australia, 06--11 Aug 2017. PMLR.
\newblock URL \url{http://proceedings.mlr.press/v70/sakr17a.html}.

\bibitem[Sheng et~al.(2018)Sheng, Feng, Zhuo, Zhang, Shen, and
  Aleksic]{Sheng2018}
Sheng, T., Feng, C., Zhuo, S., Zhang, X., Shen, L., and Aleksic, M.
\newblock {A Quantization-Friendly Separable Convolution for MobileNets}.
\newblock mar 2018.
\newblock \doi{10.1109/EMC2.2018.00011}.
\newblock URL \url{http://arxiv.org/abs/1803.08607
  http://dx.doi.org/10.1109/EMC2.2018.00011}.

\bibitem[Szegedy et~al.(2014)Szegedy, Liu, Jia, Sermanet, Reed, Anguelov,
  Erhan, Vanhoucke, and Rabinovich]{Inception_v1}
Szegedy, C., Liu, W., Jia, Y., Sermanet, P., Reed, S.~E., Anguelov, D., Erhan,
  D., Vanhoucke, V., and Rabinovich, A.
\newblock Going deeper with convolutions.
\newblock \emph{CoRR}, abs/1409.4842, 2014.
\newblock URL \url{http://arxiv.org/abs/1409.4842}.

\bibitem[Szegedy et~al.(2015)Szegedy, Vanhoucke, Ioffe, Shlens, and
  Wojna]{Szegedy2015}
Szegedy, C., Vanhoucke, V., Ioffe, S., Shlens, J., and Wojna, Z.
\newblock Rethinking the inception architecture for computer vision.
\newblock \emph{CoRR}, abs/1512.00567, 2015.
\newblock URL \url{http://arxiv.org/abs/1512.00567}.

\bibitem[TF-slim()]{TFSilm}
TF-slim, G.
\newblock Tensorflow slim models.
\newblock URL
  \url{https://github.com/tensorflow/models/tree/master/research/slim}.

\bibitem[Vanhoucke et~al.(2011)Vanhoucke, Senior, and Mao]{Vanhoucke2011}
Vanhoucke, V., Senior, A., and Mao, M.
\newblock {Improving the speed of neural networks on CPUs}.
\newblock Technical report, 2011.
\newblock URL \url{http://research.google.com/pubs/archive/37631.pdf}.

\bibitem[Zhou et~al.(2017)Zhou, Yao, Guo, Xu, and Chen]{Zhou2017}
Zhou, A., Yao, A., Guo, Y., Xu, L., and Chen, Y.
\newblock {Incremental Network Quantization: Towards Lossless CNNs with
  Low-Precision Weights}.
\newblock feb 2017.
\newblock URL \url{https://arxiv.org/abs/1702.03044}.

\bibitem[Zhou et~al.(2018)Zhou, Yao, Wang, and Chen]{Zhou_2018_CVPR}
Zhou, A., Yao, A., Wang, K., and Chen, Y.
\newblock Explicit loss-error-aware quantization for low-bit deep neural
  networks.
\newblock In \emph{The IEEE Conference on Computer Vision and Pattern
  Recognition (CVPR)}, June 2018.

\end{thebibliography}
\bibliographystyle{icml2019}

\end{document}